\documentclass[conference]{IEEEtran}
\IEEEoverridecommandlockouts

\usepackage{cite}
\usepackage{amsmath,amssymb,amsfonts}
\usepackage{algorithmic}
\usepackage{graphicx}
\usepackage{textcomp}
\usepackage{xcolor}

\usepackage{siunitx}
\usepackage{multirow}
\usepackage{booktabs}

\def\BibTeX{{\rm B\kern-.05em{\sc i\kern-.025em b}\kern-.08em
    T\kern-.1667em\lower.7ex\hbox{E}\kern-.125emX}}
    
\begin{document}

\title{Wafer Map Defect Patterns Semi-Supervised Classification Using Latent Vector Representation\\
}

\author{\IEEEauthorblockN{Qiyu Wei}
\IEEEauthorblockA{\textit{School of Microelectronics} \\
\textit{Shanghai University (SHU)}\\
Shanghai, China \\
qywei@shu.edu.cn}
\and
\IEEEauthorblockN{Wei Zhao$^{*}$}
\IEEEauthorblockA{\textit{School of Microelectronics} \\
\textit{Shanghai University (SHU)}\\
Shanghai, China \\
zw1315753898@shu.edu.cn}
\and
\IEEEauthorblockN{Xiaoyan Zheng}
\IEEEauthorblockA{\textit{School of Microelectronics} \\
\textit{Shanghai University (SHU)}\\
Shanghai, China \\
zxy20000421@shu.edu.cn}
\and
\IEEEauthorblockN{ Zeng Zeng$^{*}$}
\IEEEauthorblockA{\textit{School of Microelectronics} \\
\textit{Shanghai University (SHU)}\\
Shanghai, China \\
zengz@shu.edu.cn}
\thanks{ $^*$Corresponding author}}

\maketitle

\begin{abstract}

As the globalization of semiconductor design and manufacturing processes continues, the demand for defect detection during integrated circuit fabrication stages is becoming increasingly critical, playing a significant role in enhancing the yield of semiconductor products. Traditional wafer map defect pattern detection methods involve manual inspection using electron microscopes to collect sample images, which are then assessed by experts for defects. This approach is labor-intensive and inefficient. Consequently, there is a pressing need to develop a model capable of automatically detecting defects as an alternative to manual operations.
In this paper, we propose a method that initially employs a pre-trained VAE model to obtain the fault distribution information of the wafer map. This information serves as guidance, combined with the original image set for semi-supervised model training. During the semi-supervised training, we utilize a teacher-student network for iterative learning. The model presented in this paper is validated on the benchmark dataset WM-811K wafer dataset. The experimental results demonstrate superior classification accuracy and detection performance compared to state-of-the-art models, fulfilling the requirements for industrial applications. Compared to the original architecture, we have achieved significant performance improvement.

\end{abstract}

\begin{IEEEkeywords}
Defect detection; semi-supervised learning; unsupervised learning; variation autoencoder
\end{IEEEkeywords}

\section{Introduction}

Amid the accelerated advancement of industrial intelligence manufacturing and electronic information technology, integrated circuits are increasingly playing an essential role in the contemporary high-end chip domain. Although semiconductor foundries have achieved a high degree of manufacturing automation, the progress of advanced processes lags behind the rapid development of the industry. A high proportion and intensity of manual labor result in insufficient stability and accuracy.

Owing to the unique nature of the semiconductor industry, specialized knowledge is required to identify potential failures, often necessitating expert involvement. However, manual classification methods significantly diminish the efficiency of wafer map defect pattern analysis and resolution, not to mention the additional issues stemming from the instability of accuracy. Currently, semiconductor foundries face major challenges in defect identification and classification, circuit dimension measurement, and yield testing pattern anomaly recognition, such as excessive dependence on manual labor, low data collection utilization rates, and overall inefficiency. On-site engineers need a vast reservoir of expertise and experience, yet unified measurement standards are difficult to establish, resulting in considerable latency. With the evolution of industrial big data and artificial intelligence technologies, it is anticipated that real-time defect detection in semiconductor chips (wafers) can be achieved through image analysis, thus reducing costs and enhancing yields\cite{Xiaolei2016Wafer,2022Semi}.

Deep convolutional neural networks have achieved remarkable results in numerous fields, largely due to the availability of extensive, high-quality labeled datasets. However, the strong data confidentiality, real-time demands, and specialized nature of the semiconductor domain result in high costs and time expenditures for manual labeling\cite{2022SYSTEMS}. Particularly, disparities between production lines and short product update cycles often render timely data labeling unfeasible\cite{nakazawa2018wafer}. To address this issue, many high-performance semi-supervised image classification models have been developed, utilizing network semi-supervised datasets to achieve exceptional performance in tasks such as image classification and fine-grained recognition\cite{li2020dividemix,kong2019recycling}. However, weakly supervised learning labels possess certain flaws. Firstly, non-visual, missing, and irrelevant labels generate noise, significantly impacting model training. Secondly, weakly supervised\cite{zhou2018brief} network datasets typically adhere to Zipf's law, containing numerous long-tail labels, causing models to perform well solely on the most prominent labels. Lastly, inherent distribution characteristics of image datasets have not been adequately exploited.

To address these challenges, we explore a semiconductor defect detection method that combines a VAE model and a semi-supervised approach based on minimal data annotation. We propose a large-scale convolutional neural network-based semi-supervised learning method utilizing a teacher-student architecture and a VAE model. With labeled data, we simultaneously train the teacher and VAE models; after obtaining the teacher model, we sample unlabeled data, and the intermediate layer representing image feature distribution within the VAE is extracted as supplementary information incorporated into the teacher-student architecture to train the student model.

To sum up, the contribution of this work lies in three folds:
\begin{itemize}
\vspace{-0.1in}
\setlength{\itemsep}{0pt}
\setlength{\parsep}{0pt}
\setlength{\parskip}{0pt}
\item  We propose a novel approach for exploring fault distribution in wafer maps, employing a Variational Autoencoder (VAE) to obtain fault data distribution information from the dataset.
\item By utilizing a semi-supervised teacher-student network, we effectively leverage a large volume of unlabeled data, thereby enhancing data utilization and recognition performance.
\item Experimental results demonstrate the remarkable performance of this latent vector based semi-supervised learning method in semiconductor image recognition tasks.
\end{itemize}

\section{Related work}

\subsection{Image classification.}

Image classification is a fundamental task in computer vision, which involves assigning a label to an image from a set of predefined categories. Numerous methods have been proposed for image classification, ranging from traditional handcrafted feature-based methods to deep learning-based approaches. 
In actual industrial production, the considerable variability in object shape, size, texture, color, background, layout, and imaging illumination renders defect classification in complex environments a formidable task. Owing to the robust feature extraction capabilities of Convolutional Neural Networks (CNNs)\cite{lecun2015deep}, employing CNN-based classification networks has become the most prevalent approach for surface defect classification. Typically, the feature extraction component of a CNN classification network consists of cascading convolutional layers and pooling layers, followed by fully connected layers (or average pooling layers) and a softmax structure for classification. Generally, existing surface defect classification networks often adopt readily available network structures in computer vision, including AlexNet\cite{krizhevsky2017imagenet}, VGG\cite{simonyan2014very}, GoogLeNet\cite{szegedy2015going}, ResNet\cite{he2016deep}, DenseNet\cite{huang2017densely}, SENet\cite{hu2018squeeze}, ShuffleNet\cite{zhang2018shufflenet}, and MobileNet\cite{howard2017mobilenets}, among others. Alternatively, they may construct simplified network structures tailored to specific problems, wherein a test image is inputted into the classification network, which then outputs the image's category and the associated confidence level.
Despite significant advancements in image classification, several challenges persist, including addressing high computational costs and enhancing the model's stringent requirements for image annotation.

\subsection{Unsupervised learning.} 

At present, the most commonly used unsupervised learning models for surface defect detection are those based on normal sample learning. As these models only require normal (defect-free) samples for network training, the approach is often referred to as one-class learning. Networks trained on normal samples exclusively accept normal (non-defective) samples, endowing them with a strong ability to reconstruct and discriminate normal sample distributions. Consequently, when a network input contains defects, it often generates results distinct from those of normal samples. Compared to supervised learning models, this method can detect deviations from expected patterns or previously unseen patterns, which may be considered defects or anomalies. Based on the differences in processing spaces, this paper categorizes defect detection methods into two types: image-space-based and feature-space-based. Typically, the network models employed in this approach are autoencoders (AE)\cite{liou2014autoencoder}, variation autoencoders (VAE)\cite{doersch2016tutorial} and Generative Adversarial Networks (GANs)\cite{goodfellow2020generative}.

\subsection{Semi-supervised learning.}

Relative to fully supervised and unsupervised methods, semi-supervised methods are currently less frequently applied in surface defect detection. Semi-supervised learning is a technique in machine learning that leverages both labeled and unlabeled data for training models. In contrast to supervised learning, where models are trained only on labeled data, semi-supervised learning uses a combination of labeled and unlabeled data to improve model performance. 

Typically, semi-supervised approaches involve using image-level class annotations (weak labels) to achieve segmentation/localization-level detection results. Marino et al. \cite{marino2019weakly} employed a semi-supervised learning method based on peak response maps (PRMs) \cite{zhou2018weakly} to classify, locate, and segment potato surface defects, thereby automating quality control tasks. Mayr et al. \cite{mayr2019weakly} modified the original ResNet50\cite{he2016deep} classification network by removing the original fully connected and average pooling layers and adding two 1x1 convolutions to obtain defect response feature maps, thus enabling preliminary crack defect detection on solar panels using only image labels. Niu et al. \cite{niu2019defectgan} proposed a semi-supervised learning defect detection method based on GANs. By using CycleGAN\cite{zhu2017unpaired} to transform input test images into their corresponding defect-free images and comparing the differences between input images and generated defect-free images, surface defect detection is achieved. Semi-supervised learning typically uses a large amount of unlabeled data and a small portion of labeled data for surface defect detection model training. He et al. \cite{di2019surface} developed a semi-supervised GAN-based approach \cite{odena2016semi} for steel surface defect classification, employing a CAE-based encoder in the designed CAEGAN defect detection network and feeding it into a softmax layer to form the discriminator. The discriminator does not predict a binary classification of the input image's authenticity directly but predicts N+1 classes, where N represents the number of defect types, and the additional class denotes whether the input image originates from a real dataset or a generator. He et al. \cite{he2019semi} proposed a multi-training semi-supervised learning method for steel surface defect classification, using cDCGAN\cite{mustapha2022conditional} to generate a large number of unlabeled samples. To utilize unlabeled samples, the model introduces a multi-training fusion algorithm based on cDCGAN and ResNet-18 for unlabeled sample class label prediction. Predicted samples with assigned class labels are added to the training set for further training. This iterative process is repeated to gradually optimize the model. Extensive experiments on the NEU-CLS defect dataset \cite{he2019end} demonstrate that even with limited original samples, this method is highly effective for defect classification. Gao et al. \cite{gao2020semi} also proposed a semi-supervised learning method for classifying steel surface defects using convolutional neural networks, improving the performance of classification CNNs through the adoption of pseudo-labels. Currently, semi-supervised approaches are mostly utilized for defect classification or identification tasks and have not been widely applied to localization and segmentation tasks.

In existing semi-supervised learning methods, such as self-training, co-training, and tri-training, one issue arises: the noise level in the training dataset gradually increases during iteration. This problem can be attributed to two factors: static annotation thresholds and the uncertain timing of stopping example annotation iteration. To address these issues, researchers have proposed methods based on teacher-student networks. Similar methods were previously suggested by meta\cite{yalniz2019billion} and \cite{bhalgat2019teacher}, but due to data input issues, they were unable to effectively extract features.
\section{Proposed Methodology}

\begin{figure*}[!htbp]
\centering 
\vspace{-0.2in}
\includegraphics[width = 0.92\textwidth]{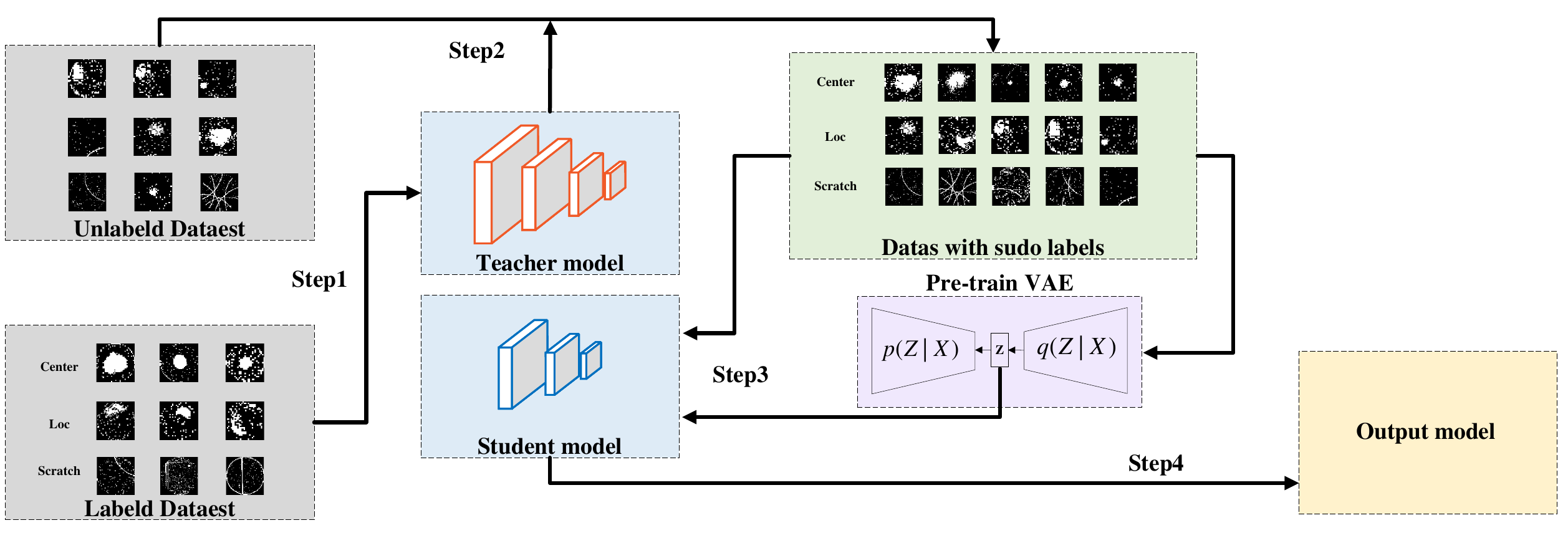}
\caption{Illustration of our approach: (1) Train an initial teacher model on the labeled dataset; (2) For each class/label pair, use the teacher model to label and score unmarked images, selecting the top-K images for each label category to construct new training data; (3) Train the student model on the newly constructed data, incorporating the VAE's latent vector; (4) Fine-tune the pre-trained student model on the initial labeled dataset to avoid potential labeling errors.}
\vspace{-0.15in}
\label{fig_overall}
\end{figure*}



\subsection{Global feature extraction via Pre-trained VAE}

Neural networks are commonly understood as approximations of the functions we want to model. However, they can also be seen as a data structure for storing information. Suppose we have a neural network composed of several deconvolutional layers. We set the input as a unit vector and train the network to minimize the mean squared error between it and the target image. In this way, the "data" of the image is contained within the current parameters of the neural network, which is the basic idea behind the Autoencoder (AE) network.
In an autoencoder, we introduce a component that automatically encodes the original image into a vector. The aforementioned deconvolutional layers can then "decode" these vectors back into the original images. However, we aim to build a generative model rather than just a "memory" of the fuzzy structure of image data. Besides encoding latent vectors from existing images as described earlier, we don't know how to create these vectors, and thus we cannot generate any images from scratch. Here's a simple solution: We add a constraint to the encoding network, forcing the latent vectors it generates to approximately follow a unit Gaussian distribution. This constraint distinguishes the Variational Autoencoder (VAE) from the standard autoencoder. The encoding latent vector is replaced by a continuous variable $Z$, this distribution can be expressed as:

\begin{equation}
\label{equ:1}
P(x)=\int_{z} P(z) P(x \mid z) d z,
\end{equation} 
where $z \sim N(0,1), \quad x \mid z \sim N(\mu(z), \sigma(z))$.

Now, generating new images becomes easy: We simply sample a latent vector from the unit Gaussian distribution and pass it to the decoder. The latent vector in the VAE, which follows a unit Gaussian distribution, is the global feature representation of wafer images that we need. To implement this functionality, we only need to pre-trained a VAE network on the existing image dataset. Afterward, each input image will yield a latent vector that represents the global features of the wafer image.

\subsection{Teacher Student Network}

To fully exploit unsupervised data, we employ a teacher-student interactive learning scheme, where the student network is optimized using pseudo-labels generated by the teacher network, and the teacher network is updated through the gradual transfer of weights from the continuously learning student network. During the interaction of teacher-student networks, both models can mutually enhance and consistently improve detection accuracy. The enhancement of detection accuracy implies that the teacher network can generate more accurate and stable pseudo-labels, which, compared to current work, we find to be crucial in significantly improving algorithm performance. On the other hand, we also regard the teacher network as an ensemble of student models at different time stages, which is consistent with our observation that the teacher network's accuracy consistently surpasses that of the student network. To address the issue of unsupervised data lacking labels, we adopt the pseudo-labeling method to train the student network using unsupervised data.

The distinction between the pseudo-labeling method and the consistency regularization method lies in the fact that consistency regularization typically relies on consistency constraints of abundant data transformations. In contrast, the pseudo-labeling method depends on high-confidence pseudo-labels, which can be added as labeled data to the training dataset. This follows the principles of current successful semi-supervised learning image classification tasks. Similar to classification-based methods, to avoid the persistent interference of noisy pseudo-labels, we first set a confidence threshold for the predicted classification results to filter out low-confidence labels.

Furthermore, noisy pseudo-labels can affect the pseudo-label-generating model (teacher network). Therefore, we separate the teacher network and the student network. For increased accuracy, only the learnable weights of the student network can be updated through backpropagation after obtaining pseudo-labels from the teacher network.

\subsection{Network Architecture}

In this study, we propose a VAE-based latent feature extraction method to detect fault regions and employ a teacher-student network to handle unlabeled data to enhance the performance of the classification model, the whole architecture can be seen in Figure \ref{fig_overall}. The following are the key steps we adopt: Firstly, we pre-train a VAE model using the wafer dataset. This model can automatically learn and identify fault distribution features in images without relying on manually annotated data, with the latent vector in the intermediate layer being the feature data we need. Next, we construct the initial teacher model, which is trained on a limited amount of labeled data to attain sufficient performance to guide the student model initially.
After building the teacher model, we label and score the unlabeled images. This process typically uses the teacher model's output as a reference to generate pseudo-labels for the unlabeled images. Subsequently, we select the top-K images with the highest confidence from the pseudo-labeled images and combine them with the original labeled data to create a new training dataset. Utilizing this new training dataset, along with the latent vector obtained after passing the dataset through the VAE, we train the student model. The student model learns by observing the teacher model's behavior during this process.
Lastly, after training the student model, we fine-tune it. This step can further enhance the model's performance, making it better suited for practical tasks. In summary, by adopting our proposed method based on unsupervised semantic segmentation, constructing a teacher model, labeling, training a student model, and fine-tuning, we successfully improve the model's performance when dealing with unlabeled data.

\section{Experiments}

\subsection{Dataset and Evaluation Protoco}

\begin{figure}[!htbp]
\centering 
\includegraphics[width = 0.43
\textwidth]{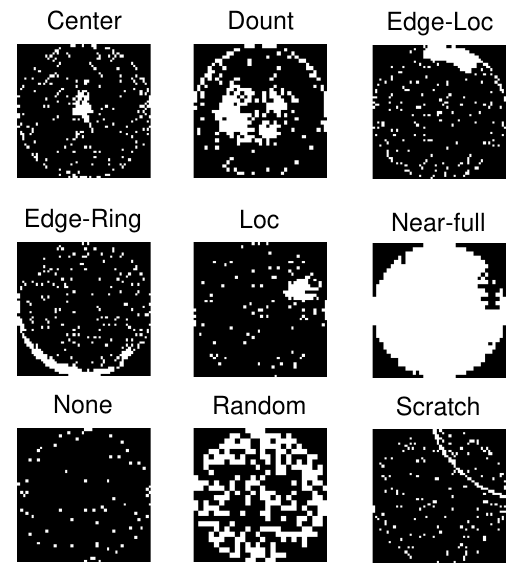}
\caption{ Examples of wafer map failure patterns }
\label{fig1}
\end{figure}

{\bf Dataset.} To demonstrate the superiority of the model algorithm presented in this paper, we have selected the WM-811K semiconductor dataset for experimentation. This extensive wafer database, originating from a Kaggle competition, encompasses 811,457 wafer images, along with additional information such as wafer core dimensions, batch numbers, and wafer indices. The manually labeled dataset consists of 172,950 images, with a total of nine labels (0-8), which can be seen in Figure \ref{fig1}; label 8 represents defect-free, normal wafers, accounting for $85\%$ of the entire dataset, while labels 0-7 correspond to defective wafer data.


\textbf{Evaluation Metrics.} In this paper, we employ four widely used evaluation metrics, namely Precision, Recall, F1 score, and Accuracy, to assess the results of our experiments. These metrics are commonly used to evaluate recognition accuracy in the context of imbalanced data.
These indices are calculated as follows: 
\begin{equation}
\text { Precision }  =\frac{\mathrm{TP}}{\mathrm{TP}+\mathrm{FP}},
\end{equation}
\begin{equation}
\text { Recall }  =\frac{\mathrm{TP}}{\mathrm{TP}+\mathrm{FN}},
\end{equation}
\begin{equation}
\text { F1-score }=\frac{2 \cdot \text { Precision } \cdot \text { Recall }}{\text { Precision }+\text { Recall }},
\end{equation}
\begin{equation}
\text { Accuracy }  =\frac{\mathrm{TP}+\mathrm{TN}}{\mathrm{TP}+\mathrm{FN}+\mathrm{TN}+\mathrm{FP}},
\end{equation}
where the notation TP, FN, TN, and FP are used to represent true positive samples, false negative samples, true negative samples, and false positive samples, respectively.

\subsection{Implementation Details.}

The experiments conducted in this study were based on the CentOS 7 system, utilizing Python 3.8 and the PyTorch v1.9.1 framework. The hardware specifications included an NVIDIA GeForce RTX A5000 with 24GB of video memory. The Adam optimizer \cite{kingma2014adam} was employed for network optimization, with an initial learning rate (LR) of 0.002, momentum parameters $\beta 1 = 0.9$ and $\beta 2 = 0.99$, and a training epoch setting of 100.

For both the teacher and student models, we employ Residual Networks (ResNet-50) as the backbone architecture, the details of the network structure used in our experiments are shown in the table.


\begin{table}[!t]
\caption{Details of Resnet50.}
\label{tab:resnet50_details}
\centering
\begin{tabular}{@{}ccccc@{}}
\toprule
                          & layer name & output size & 50-layer                                                             & Repeats \\ \midrule
                          & conv1      & 112*112     & 7*7,64                                                              & - \\ \cmidrule(l){2-5} 
                          & conv2      & 56*56       & \begin{tabular}[c]{@{}c@{}}1*1,64\\ 3*3,64\\ 1*1,256\end{tabular}    & 3 \\ \cmidrule(l){2-5} 
\multirow{6}{*}{ResNet50} & conv3      & 28*28       & \begin{tabular}[c]{@{}c@{}}1*1,128\\ 3*3,128\\ 1*1,512\end{tabular}  & 3 \\ \cmidrule(l){2-5} 
                          & conv4      & 14*14       & \begin{tabular}[c]{@{}c@{}}1*1,256\\ 3*3,256\\ 1*1,1024\end{tabular} & 6 \\ \cmidrule(l){2-5} 
                          & conv5      & 7*7         & \begin{tabular}[c]{@{}c@{}}1*1,512\\ 3*3,512\\ 1*1,2048\end{tabular} & 3 \\ \cmidrule(l){2-5} 
                          &            & 1*1         & average pool                                                         & - \\ \bottomrule
\end{tabular}
\end{table}

ResNet is divided into five stages, where Stage 0 has a relatively simple structure and can be considered as preprocessing for the input. The following four stages consist of Bottleneck structures and are relatively similar. Stage 1 contains three Bottlenecks, while the remaining three stages encompass four, six, and three Bottlenecks respectively. Each stage contains distinct information, constrained by the convolution and receptive field; earlier stages primarily process local information, whereas later stages gradually aggregate local information to obtain global information. We attempt to introduce the latent vector extracted by the VAE between the five stages, testing the effect of incorporating global distribution information at different stages.
It is worth noting that when incorporating latent vectors of varying sizes at different positions, the size of the intermediate latent vector within our VAE must be adjusted accordingly.

\noindent\textbf{Unbalanced Labeled Data.} We investigated the distribution of patterns in the dataset and observed a noticeable imbalance in the original dataset. This class imbalance can adversely affect the performance of traditional classifiers. Non-Pattern modes constitute the vast majority, while Dount and NearFull modes account for only $0.3\%$ and $0.1\%$, respectively. To address this issue, we explored oversampling and undersampling techniques, balancing the labeled data in the dataset so that the number of images for each class is approximately two thousand.

\noindent\textbf{Strategy of pseudo labels selection.} 
To maintain the quality of pseudo-labels, analyze the model's prediction probabilities for each unlabeled image. Set a confidence threshold as 0.9, and only retain images where the model's prediction probability is above this threshold. Then add the selected high-confidence pseudo-labeled images to the labeled dataset. The augmented dataset now contains a mix of true labels and pseudo-labels.

\subsection{Performance study}

\noindent\textbf{Quantitative Comparison}. To demonstrate the effectiveness of adding the latent vector as global information to the input of the classification performance, we compared the performance of models with and without latent vector in different ResNet models: our proposed method with latent vector and the standard ResNet without latent vector. Both models were trained from scratch under the same settings. The comparison results of the two models on the test set are shown in Table II.


\begin{table}[!t]
    \centering
    \setlength{\tabcolsep}{1.5mm}
    \caption{Quantitative Comparison.}
    \label{tab:1}
    \begin{tabular}{ccccc}
        \hline
                    & P     & R     & F1    & A     \\ 
        \hline
        Without VAE & 0.932 & 0.911 & 0.944 & 0.961 \\ 
        \hline
        With VAE    & 0.946 & 0.912 & 0.962 & 0.977 \\ 
        \hline
    \end{tabular}
\end{table}

From the table, we can observe that by incorporating the latent vector representing global features, the overall classification performance of the model is significantly improved. In particular, the sensitivity values of our proposed method are much higher than those of the standard method, indicating a significant improvement in the classification accuracy of wafer defects.
%
%
%

\begin{table}[!t]
    \centering
    \setlength{\tabcolsep}{1.5mm}
    \caption{Abalation Study.}
    \label{tab:2}
    \begin{tabular}{lllll}
        \hline
        & P & R & F1 & A \\ 
        \hline
        Behind Stage1 & 0.811 & 0.701 & 0.761 & 0.891 \\ 
        \hline
        Behind Stage2 & 0.946 & 0.912 & 0.962 & 0.977 \\ 
        \hline
        Behind Stage3 & 0.671 & 0.724 & 0.701 & 0.811 \\ 
        \hline
        Behind Stage4 & 0.656 & 0.681 & 0.637 & 0.737 \\ 
        \hline
    \end{tabular}
\end{table}

\noindent\textbf{Abalation study.} In order to investigate the impact of the position of the latent vector insertion within the ResNet architecture on model performance, we conducted an ablation study. As illustrated in Table III, we examined the performance of various network structures on the test set when the insertion positions were set to stages 1, 2, 3, and 4, respectively. Based on a comprehensive comparison of the results, the performance is optimal when the latent vector is incorporated after stage 2, demonstrating the most significant enhancement in model performance due to the global distribution information from the VAE. In light of these observations, we have chosen to integrate the model by inserting the latent vector after stage 2 in the ResNet architecture.

\section{Conclusions}
%
The work employs a semi-supervised learning approach to detect defects on wafer surfaces, utilizing the classic teacher-student model in the field of semi-supervised object detection, aiming to reduce the high costs associated with labeling large quantities of images. While maintaining the teacher-student model framework, improvements are made to the teacher and student models in the form of a feature fusion mechanism. By altering the serial structure of the ResNet network in the teacher-student model to a parallel structure with a VAE, the difficulty of extracting global wafer defect features is circumvented. Ablation experiments and lateral comparison experiments conducted on the W811K dataset validate the practicality and effectiveness of the improved teacher-student model from different perspectives.


\bibliographystyle{IEEEtran}
\bibliography{ref}

\end{document}